\documentclass{article} 
\usepackage{collas2025_conference,times}
\usepackage{easyReview}

\usepackage{amsfonts,amsmath,amssymb,amsthm}
\usepackage{mathtools}

\newtheorem{theorem}{Theorem}[section]

\newtheorem{lemma}[theorem]{Lemma}

\usepackage{hyperref}

\usepackage{wrapfig}

\usepackage{graphicx}

\usepackage{algorithm}
\usepackage{algpseudocode}

\usepackage{hyperref}
\hypersetup{
colorlinks=true,
linkcolor=red,
filecolor=magenta,
urlcolor=blue,
citecolor=purple,
pdftitle={SSF},
pdfpagemode=FullScreen,
}

\makeatletter
\renewcommand{\And}{\end{tabular}\hspace{11em}\begin{tabular}[t]{c}}  
\makeatother

\title{Extremely Simple Streaming Forest}

\author{Haoyin Xu \\
Department of Biomedical Engineering\\
Johns Hopkins University\\
USA \\
\texttt{hx@jhu.edu} \\
\And 
Jayanta Dey \\
Department of Biomedical Engineering\\
Johns Hopkins University\\
USA \\
\texttt{jdey4@jhmi.edu} \\
\AND
Sambit Panda\\
Department of Biomedical Engineering\\
Johns Hopkins University\\
USA \\
\texttt{spanda3@jhu.edu} \\
\And 
Joshua T. Vogelstein\\
Department of Biomedical Engineering\\
Johns Hopkins University\\
USA \\
\texttt{jovo@jhu.edu}
}

\preprintcopy 
\begin{document}

\maketitle
\begin{abstract}
Decision forests, including random forests and gradient boosting trees, remain the leading machine learning methods for many real-world data problems, especially on tabular data.
However, most of the current implementations only operate in batch mode, and therefore cannot incrementally update when more data arrive. 
Several previous works developed streaming trees and ensembles to overcome this limitation. Nonetheless, we found that those state-of-the-art algorithms suffer from a number of drawbacks, including low accuracy on some problems and high memory usage on others. 
We therefore developed an extremely simple extension of decision trees: given new data, simply update existing trees by continuing to grow them, and replace some old trees with new ones to control the total number of trees. 
In a benchmark suite containing 72 classification problems (the OpenML-CC18 data suite), we illustrate that our approach, \textit{Extremely Simple Streaming Forest} (XForest), does not suffer from either of the aforementioned limitations. On those datasets, we also demonstrate that our approach often performs as well as, and sometimes even better than, conventional batch decision forest algorithms. 
With a \textit{zero-added-node} approach, XForest-Zero, we also further extend existing splits to new tasks, and this very efficient method only requires inference time.
Thus, XForests establish a simple standard for streaming trees and forests that could readily be applied to many real-world problems.
\end{abstract}

\section{Introduction}
\label{introduction}
In recent decades, machine learning methods facilitate the utilization of modern data and have made significant scientific progress in health care, technology, commerce, and more \citep{jordan_machine_2015}. 
Among these methods, random forests, hereafter referred to as random forests (RFs), and gradient boosting trees (GBs) are the leading strategies for tasks like classifications, outperforming all others on real-world datasets and machine learning competitions \citep{breiman_random_2001, friedman_greedy_2001, caruana_empirical_2006, caruana_empirical_2008, fernandez-delgado_we_2014, chen_xgboost_2016}.
However, training forests with continuous inputs poses new challenges for these estimators.
Larger sample sizes often require overwhelmingly more computational time and space, and different scenarios of streaming inputs need flexible strategies for model updates \citep{liu_isolation_2008, abdulsalam_streaming_2007}. This condition requires that, for batch tree estimators like RFs, all available sample data must be stored and refitted after each update \citep{amit_shape_1997, breiman_random_2001}. Even with enough computational resources to do so, out-of-distribution (OOD) problems could still undermine the validity of older data \citep{geisa_towards_2021}. 

By incrementally updating the estimators with continuous batches of new training samples, streaming trees can continuously optimize the tree structures without storing old data \citep{domingos_mining_2000, bifet_adaptive_2009, lakshminarayanan_mondrian_2014, ben-haim_streaming_2010}. Furthermore, in forest ensembles, older trees could be pruned to keep up with the current data distribution. Attempts have been made to adapt decision trees (DTs) to streaming tasks, including Hoeffding trees (HTs) which use statistical bounds to update node splits, and Mondrian forests (MFs) which create random partitions \citep{domingos_mining_2000, lakshminarayanan_mondrian_2014}. These two state-of-the-art algorithms have been tested in many synthetic and real-world scenarios \citep{pfahringer_new_2007, gomes_machine_2019, lakshminarayanan_mondrian_2016, khannouz_benchmark_2020}. Nonetheless, we find them performing poorly in certain tasks and using excessive memories on large datasets, which significantly restrict their applications.

In this paper, we explore an extremely simple incremental update method to extend a fitted decision tree and introduce our streaming tree implementations: \textit{Extremely Simple Streaming Tree} (XTree) and \textit{Extremely Simple Streaming Forest} (XForest).
After the initial fitting, an XTree takes in every new batch of training samples and splits its leaf nodes when satisfying the given criteria, such as a sample minimum. An XForest ensembles XTrees with bootstrapping and replaces old trees with new ones to control the total number of estimators.
On three standard datasets with varying complexities, the performance of streaming and batch estimators are compared, and we illustrate the consistently high performance of XForests without the aforementioned limitations of HTs and MFs. Then in the OpenML-CC18 benchmark suite,
which contains 72 diverse classification tasks, we demonstrate that XForests often achieve accuracy similar to, or even better than, those of batch mode RFs and GBs \citep{vanschoren_openml_2013, bischl_openml_2019}.

We also model after forest honesty and apply a \textit{zero-added-node} approach (XForest-Zero) by refitting leaf nodes with training samples from another task \citep{athey_generalized_2018}. This very efficient method only needs inference time to transfer existing splits to new tests, and it does not consume any additional resources. We illustrate the viability of XForest-Zeros by applying MNIST splits to Fashion-MNIST data, and they also achieve performance comparative to batch mode RFs \citep{lecun_gradient-based_1998, xiao_fashion-mnist_2017}.
Overall, we believe our simple approach to streaming trees could be readily applied to many real-world problems, including OOD problems and continual learning \citep{geisa_towards_2021, van_de_ven_three_2019}.

\section{Methods} 
\label{methods}

\subsection{XTrees and XForests}

\begin{minipage}[t]{0.49\columnwidth}
\begin{algorithm}[H]
\caption{Incrementally update an XTree with a batch of training samples.}
\label{alg:XTree}
\begin{algorithmic}[1]
\Require 
\Statex (1) $T$ \Comment{current tree}
\Statex (2) $\mathcal{D}_n = (\mathbf{x},\mathbf{y}) \in \mathbb{R}^{n \times p} \times \{1,\ldots, K\}^n$ \Comment{samples}
\Statex (3) split criteria \Comment{e.g., minimum node size}
\Ensure
$T$ \Comment{updated tree}
\Function{XTree.update\_tree}{$T,\mathbf{x},\mathbf{y}$}
\For{each $i \in [n]$}
\State find the leaf node $x_i$ would fall into
\State label that leaf node as a ``false root''
\State update index set for this leaf node to include $i$
\EndFor
\For{each ``false root''}
\State split the node if satisfying split criteria
\For{both child nodes}
\State mark as ``false roots''
\EndFor 
\State mark the node as internal or leaf
\EndFor
\State \Return $T$
\EndFunction 
\end{algorithmic}
\end{algorithm}
\end{minipage}
\hfill
\begin{minipage}[t]{0.5\columnwidth}
\begin{algorithm}[H]
\caption{Incrementally update an XForest with a batch of training samples.}
\label{alg:XForest}
\begin{algorithmic}[1]
\Require
\Statex (1) $\mathbf{T}$ \Comment{current forest}
\Statex (2) $\mathcal{D}_n = (\mathbf{x},\mathbf{y}) \in \mathbb{R}^{n \times p} \times \{1,\ldots, K\}^n$ \Comment{samples}
\Statex (3) split criteria \Comment{e.g., minimum node size}
\Statex (4) $r$ \Comment{number of trees to replace}
\Ensure
$\mathbf{T}$ \Comment{updated forest}
\Function{XForest.update\_forest}{$\mathbf{T},\mathbf{x},\mathbf{y},r$}
\For{each tree in the forest}
\State resample ($\mathbf{x},\mathbf{y}$) with bootstrapping
\State limit the number of features used for splitting
\State update the tree with resampled data 
\EndFor
\State record $b$, the number of batches seen
\If{$b$ is greater than 1}
with $1/b$ probability:
\State find the accuracy of trees with respect to the current batch
\State sort the indices of worst performing trees
\For{each $i \in [r]$}
\State resample ($\mathbf{x},\mathbf{y}$) with bootstrapping
\State create a new tree with resampled data
\State replace the $i$-th worst tree with the new tree 
\EndFor
\EndIf
\State \Return $\mathbf{T}$
\EndFunction
\end{algorithmic}
\end{algorithm}
\end{minipage}

\begin{figure}[!htb]
\begin{minipage}[c]{0.5\columnwidth}
    \includegraphics[width=\columnwidth]{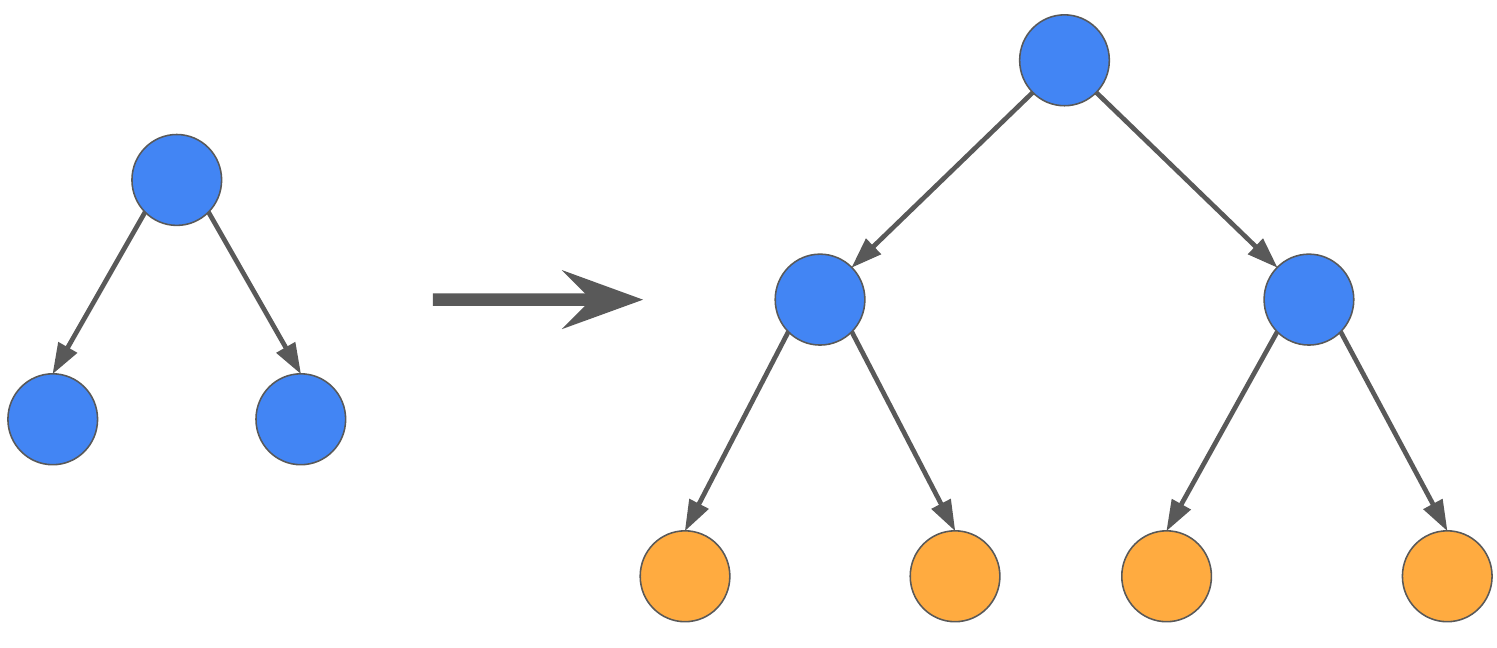}
\end{minipage}\hfill
\begin{minipage}[c]{0.4\columnwidth}
    \caption{Extremely Simple Streaming Tree structures after the first batch \textbf{(left)} and the second batch of samples \textbf{(right)}. After satisfying the splitting criteria, the leaf nodes \textbf{(left)} are splitted with respect to the second data batch. The orange circles represent the added nodes. The internal splits are never modified, preserving earlier partitions of the feature space.
    }
\label{fig:demo}
\end{minipage}
\end{figure}

XTrees are based on a fork
of the scikit-learn python package (BSD-3-Clause), which we customized by adding a new \texttt{partial\_fit} function to the \texttt{DecisionTreeClassifier} for extending fitted trees \citep{pedregosa_scikit-learn_2011}. Thus, an XTree is initialized almost in the same way as a DT, where the tree is fitted to the first batch of training samples. The only difference is the input of predefined class labels, as the first batch might not include all possible classes. 
For each incremental update afterwards (Algorithm \ref{alg:XTree}), the XTree takes in one batch of training samples: $\mathcal{D}_n = (\mathbf{x},\mathbf{y}) \in \mathbb{R}^{n \times p} \times \{1,\ldots, K\}^n$. 
Then for every training sample $x_i$ in the batch, the algorithm locates the leaf node it would fall into and associates the node with the sample. All the leaf nodes found in such a manner are marked as ``false roots.'' 
These ``false roots'' are then split and extended if satisfying the prespecified splitting criteria (e.g. a minimum node sample size of two). 
If the splitting occurs, both child nodes would be recursively marked as ``false roots'' and associated with their separate training samples, just like how batch splitting works in DTs. If not fulfilling the splitting criteria, the ``false roots'' would become the new leaf nodes. Figure \ref{fig:demo} shows an example of how XTree structures evolve. 

An XForest
is initialized as a collection of XTrees (100 by default) and modeled after scikit-learn's RF implementation: \texttt{RandomForestClassifier} \citep{pedregosa_scikit-learn_2011}. As in Algorithm \ref{alg:XForest}, it incrementally updates the trees by bootstrapping the training batches and limiting the number of features selected per split (``max\_features''). The default ``max\_features'' is defined as $p=\sqrt{d}$, where $d$ is the total number of features. After each incremental update, with $1/b$ probability, where $b$ is the number of batches seen, the forest replaces its worst performing trees (one tree by default) with new trees fitted only to the current batch. 
This step would remove XTrees that are negatively affected by earlier partitions and reduce the computational space taken. All predictions are generated via majority voting, which helps improving the ensemble's consistency \citep{liaw_classification_2002, biau_consistency_2008, breiman_random_2001}. Furthermore, the algorithm uses the joblib python package (BSD-3-Clause) and takes advantage of parallel computing \citep{joblib_developers_joblibjoblib_2022}.

In all benchmarks, we incrementally updated XTrees and XForests with fixed-sized data batches (100 random samples per batch) and measured their performance on separated test sets.

\subsection{XForest-Zero}
An XForest-Zero refits the leaf nodes of an existing forest without adding new splits to the forest. The method involves two similar tasks with potentially common splitting patterns, so that the splits generated from the first task can be applied to the second. After an XForest-Zero is trained on the first task, similar to regular XForests, it receives a batch of refitting samples. Then by inferencing the decision paths, the algorithm associates these samples with the corresponding leaf nodes. Instead of splitting those leaves, the method updates the posterior probabilities according to the class labels, which is similar to forest honesty \citep{athey_generalized_2018}.

\subsection{Reference Algorithms}
For comparison with state-of-the-art streaming tree algorithms, we experimented with two existing estimators: Hoeffding trees and Mondrian forests \citep{domingos_mining_2000, lakshminarayanan_mondrian_2014, khannouz_benchmark_2020, pfahringer_new_2007, gomes_machine_2019, lakshminarayanan_mondrian_2016}. 

HTs are designed to grow tree learners incrementally using Hoeffding bounds \citep{domingos_mining_2000, hoeffding_probability_1994}. 
After observing $n$ samples within range $R$, the bound states that, with $1 - \delta$ probability, the variable's true mean is no smaller than the sample mean minus $\epsilon$, where 
\begin{equation*} \epsilon=\sqrt{\frac{R^2\ln(\delta^{-1})}{2n}} \tag{1} \end{equation*}
Thus, the algorithm's robustness is determined by specifying the $\delta$ value.
This statistical bound is independent of heuristic measures (e.g. Gini index and information gain), and the tree splits the current leaf node after observing enough samples.

When training with large datasets, HTs could take up excessive memories to account for all growing leaves \citep{domingos_mining_2000, lavanya_handwritten_2017}. Thus, resource constraints would temporarily deactivate low performing leaves and reactivate them as needed. The algorithm constantly monitors the error rates of all leaf nodes and the probabilities of samples falling into them. 
For experiments, we used \texttt{HoeffdingTreeClassifier} from the river python package (BSD-3-Clause) \citep{montiel_river_2020}. The max size of HTs was set as 1,000 MB, and the minimum sample size for splitting (``grace\_period'') was changed to two.

On the other hand, MFs utilize Mondrian processes to incrementally grow Mondrian trees, which are named after Piet Mondrian's paintings\citep{lakshminarayanan_mondrian_2014, roy_mondrian_2009}. The Mondrian processes recursively construct random and hierarchical partitions in the feature space. They assign probabilities in a consistent way, and new splits cannot intercept the existing splits \citep{lakshminarayanan_mondrian_2014}. Specifically in the one-dimensional case, the partitions are shown to be a Poisson point process \citep{roy_mondrian_2009}. The forests utilize up to three types of operations when introducing a new sample: splitting existing leaves, updating existing partitions, and creating new splits from internal nodes \citep{lakshminarayanan_mondrian_2014}.
We used \texttt{MondrianForestClassifier} from the scikit-garden python package (BSD-3-Clause) \citep{kumar_scikit-gardenscikit-garden_2017}. The total number of estimators (``n\_estimators'' = 10) was limited in some tasks to ensure enough computational space, which is further explored in Section \ref{results:stream}.

For comparison with batch tree estimators, we included batch decision trees and forests \citep{breiman_random_2001, friedman_greedy_2001}. A RF contains a collection of DTs (100 by default) and uses bootstrapping to resample the training data. A GB also implements DTs and improves performance with boosting. Each tree in the forest limits the number of features selected at each node split (``max\_features'') and tries to find the best partitions available \citep{breiman_random_2001}.
With majority voting as predictions, a RF is non-parametric and universally consistent, so it will approach Bayes optimal performance with sufficiently large sample sizes, tree depths, and number of trees \citep{liaw_classification_2002, biau_consistency_2008}.
Implementations of DT and RF were from the scikit-learn package: \texttt{DecisionTreeClassifier} and \texttt{RandomForestClassifier} \citep{pedregosa_scikit-learn_2011}. GB implementation was from the XGBoost package: \texttt{XGBClassifier} \citep{chen_xgboost_2016}. All hyperparameters were kept as default except the number of trees.

In all tasks, we incrementally updated HTs and MFs with fixed-sized data batches (100 random samples per batch). At each sample size, DTs, RFs, and GBs were trained with all available data, including current and previous batches. 

\subsection{Data}
We used the OpenML-CC18 data suite\footnote{\url{https://www.openml.org/s/99}} for experiments. It represents a collection of 72 real-world datasets organized by OpenML, functioning as a comprehensive benchmark suite \citep{vanschoren_openml_2013, bischl_openml_2019}. These datasets vary in sample size, feature space, and unique target classes.
About half of the tasks are binary classifications, and the other half are multiclass classifications with up to 50 classes. The range of total sample sizes is between 500 and 100,000, while the range of features is from a few to a few thousand \citep{bischl_openml_2019}.

\begin{wraptable}[9]{L}{7cm}
\centering
\begin{tabular}{lccc} 
\hline
Dataset & \textbf{Splice} & \textbf{Pendigits} & \textbf{CIFAR-10} \\
\hline
\# Features & $60$ & $16$ & $3,072$ \\
\hline
\# Classes & $3$ & $10$ & $10$ \\
\hline
\# Train & $2,392$ & $7,494$ & $50,000$ \\
\hline
\# Test & $798$ & $3,498$ & $10,000$ \\
\hline
\end{tabular}
\caption{Dataset attributes for three selected tasks.}
\label{table:data}
\end{wraptable}

In XTree and XForest experiments, we selected three standard datasets (Table \ref{table:data}) for more in-depth analyses: Splice-junction Gene Sequences (Splice), Pen-Based Recognition of Handwritten Digits (Pendigits), and CIFAR-10 \citep{towell_molecular_1991, alpaydin_pen-based_1998, krizhevsky_learning_2012, lavanya_handwritten_2017}.

For the Splice dataset, we randomly reserved 25\% of all genetic sequences for testing \citep{rampone_splice-junction_1998, sarkar_splice_2020}. Each sequence contains 60 nucleotide codes, which are converted into numerical features by the OpenML curation \citep{feurer_openml-python_2019}. This task tests MFs' performance on DNA data without feature engineering, which was not satisfactory in the original paper \citep{lakshminarayanan_mondrian_2014}. 
The Pendigits dataset contains handwritten images represented by 16 pixel features, which are generated through spatial resampling \citep{alimoglu_methods_1996}.
The CIFAR-10 dataset contains RGB-colored images, each high-dimensional image represented by $32 \times 32 \times 3 = 3,072$ pixel features \citep{krizhevsky_learning_2012}. We used the provided training and test sets for Pendigits and CIFAR-10, and ran each of the three tasks 10 times by randomizing the training sets. 

In XForest-Zero experiments, the pair of MNIST and Fashion-MNIST was selected due to the similarities of these two datasets. Both image sets contain $28 \times 28 = 784$ pixel features and 10 classes, which allows seamless transitions between training, refitting and testing \citep{lecun_gradient-based_1998, xiao_fashion-mnist_2017}. The samples are, however, not related across tasks, representing handwritten digits and fashion clothing, respectively. We randomized the datasets 10 times and used 5-fold cross validation to generate subsets for training (56,000 MNIST), refitting (500 Fashion-MNIST), and testing (14,000 Fashion-MNIST). Then we trained XForest-Zeros (100 trees) with the subsets and compared their performance with RFs (100 trees) trained only on the refitting samples.

Datasets were imported using the OpenML-Python package (BSD-3-Clause) \citep{feurer_openml-python_2019}. In the OpenML-CC18 tasks, We used all available samples and ran 5-fold cross validations with forests.

\subsection{Evaluation Metrics}
Classifier performance is evaluated by classification accuracy, AUROC (one-vs-rest for multi-class tasks), classifier size, maximum training space, and training wall times. 
On the three selected datasets, we use the psutil python package (BSD-3-Clause) to measure memory usage for all the repetitions
\citep{rodola_giampaolopsutil_2022}.
The training wall times calculate the fitting times for the given model without hyperparameter tuning (Appendix \ref{app:select_time}). All estimators' times are calculated by accumulating the training times of data batches, which account for batch estimators' refitting. We do not report any error bars as they are minimal. 
To accommodate the space and time constraints of MFs and HTs, the three selected tasks were run without parallelization on a Microsoft Azure 6-core (Intel Xeon E5-2690 v3) Standard\_NC6 instance with 56 GB memory and 340 GB SSD storage.
The OpenML-CC18 experiments were run with parallelization on a Microsoft Azure 4-core (Intel Xeon E5-2673 v4) Standard\_D4\_v3 instance with 16 GB memory and 100 GB SSD storage.

\section{Results}
\label{results}

\begin{figure*}[!htb]
\centering
\includegraphics[width=0.9\textwidth]{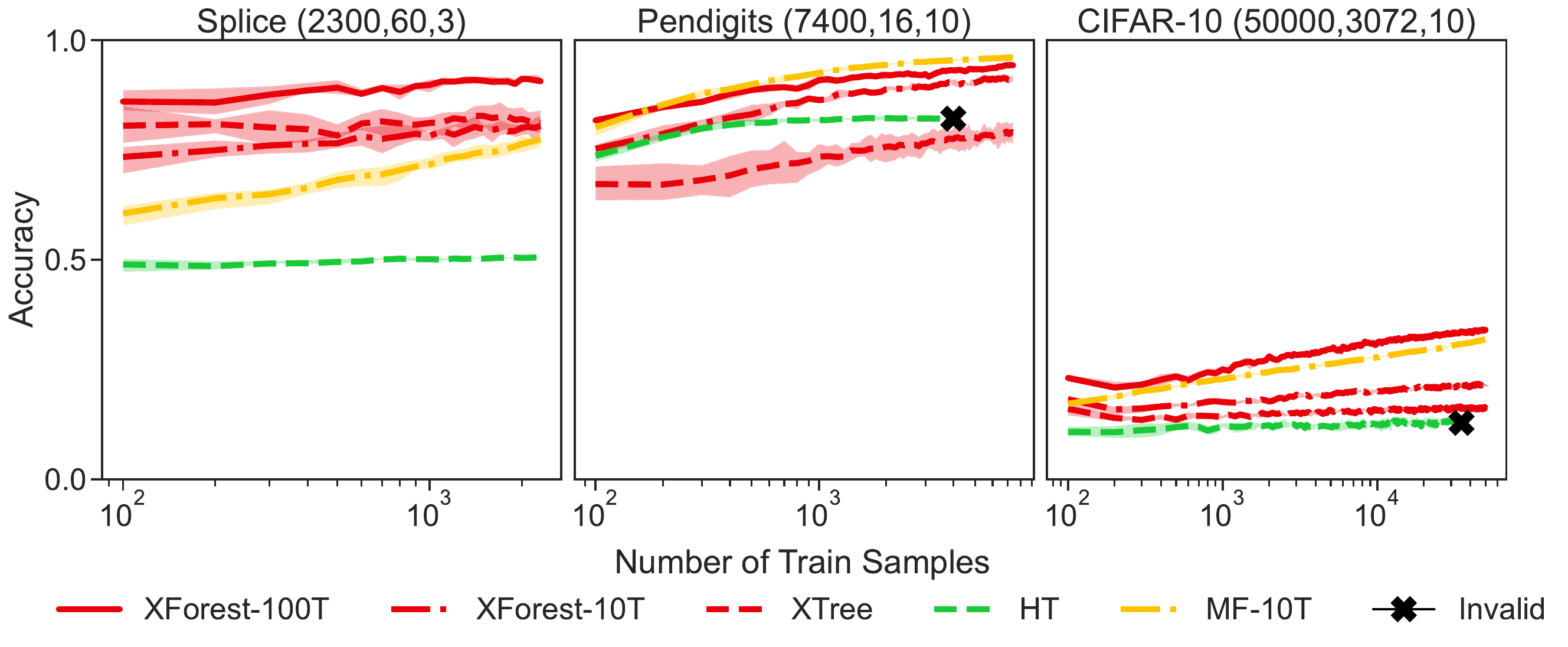}
  \caption{Classifier accuracy on three selected datasets. The panel titles include the maximum sample size, the number of features, and the number of classes.
  The classifiers include: Extremely Simple Streaming Forest with 100 trees \textbf{(XForest-100T)} and 10 trees \textbf{(XForest-10T)}, Extremely Simple Streaming Tree \textbf{(XTree)}, Hoeffding Tree \textbf{(HT)}, and Mondrian Forest with 10 trees \textbf{(MF-10T)}. 
  Each line represents averaged results from \textbf{10 randomized repetitions}. The shaded regions highlight the \textbf{25th} through \textbf{75th} percentiles. XForests perform better than all other estimators on the Splice and CIFAR-10 datasets. After certain sample sizes, the accuracy of HTs experiences significant fluctuations and becomes invalid. Those results are not shown, and the sample sizes are marked with the \textbf{cross symbol}. MFs perform slightly better than XForests in the Pendigits task, but their accuracy significantly drops in the Splice task, only surpassing that of HTs.
  }
\label{fig:select_acc_stream}
\end{figure*}

\subsection{Accuracy Comparisons with Streaming Algorithms}
\label{results:stream}
XForests consistently achieve high accuracy across all sample sizes and classification tasks, as compared to MFs and HTs. Specifically in the Splice and CIFAR-10 tasks (Figure \ref{fig:select_acc_stream}, left and right), the algorithm always performs the best among streaming classifiers. Its accuracy is also very close to the highest on the Pendigits dataset (Figure \ref{fig:select_acc_stream}, center). 
Single XTrees also outperform HTs in the Splice and CIFAR-10 tasks (Figure \ref{fig:select_acc_stream}, left and right).
These results clearly show that our streaming trees and ensembles incrementally optimize the tree structures with new information, benefiting from streaming data batches.

\begin{figure}[!htb]
\begin{minipage}[c]{0.4\columnwidth}
    \includegraphics[width=\columnwidth]{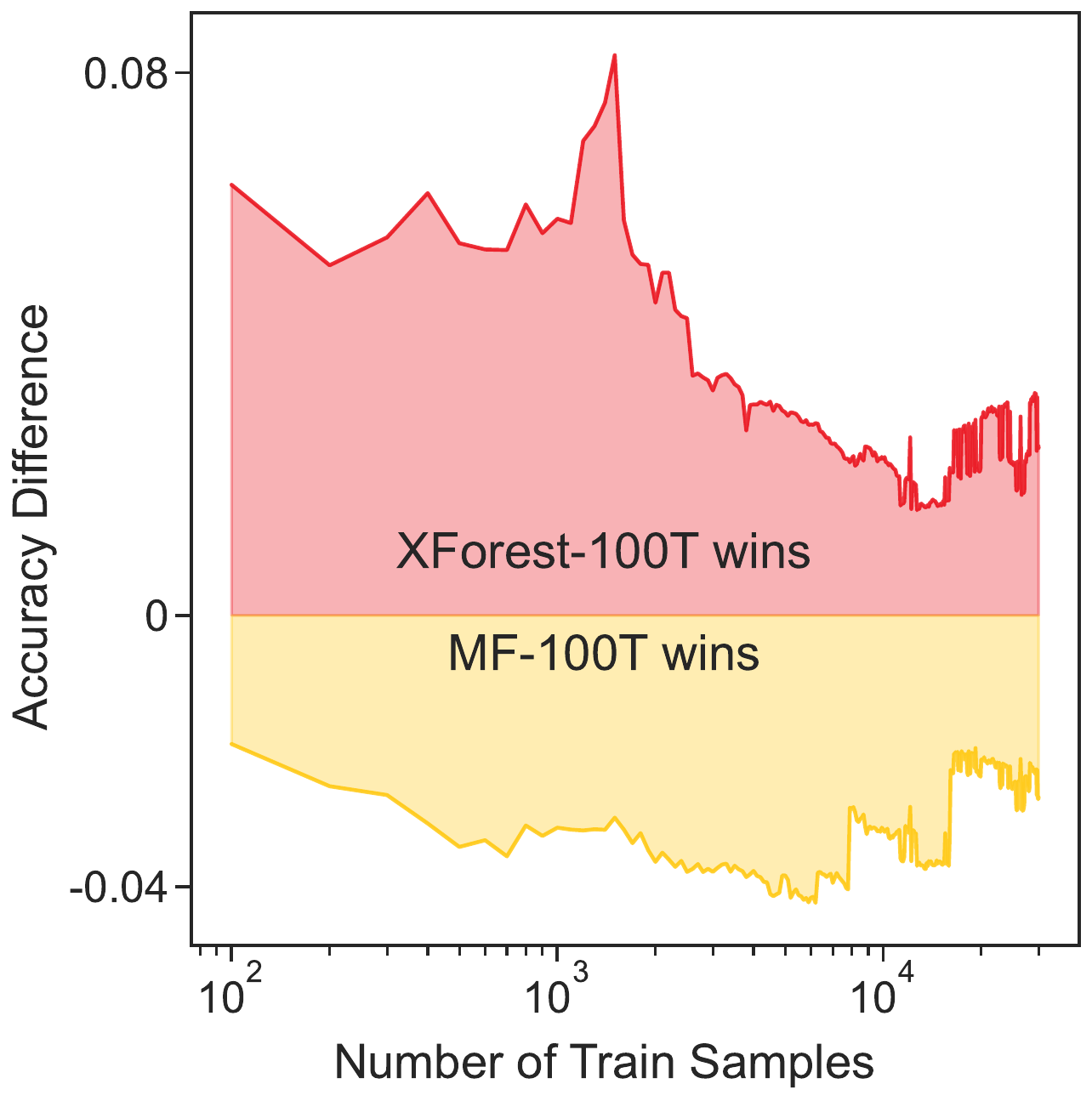}
\end{minipage}\hfill
\begin{minipage}[c]{0.55\columnwidth}
    \caption{Accuracy differences between Extremely Simple Streaming Forest with 100 trees \textbf{(XForest-100T)} and Mondrian Forest with 100 trees \textbf{(MF-100T)}. 
    The results include \textbf{67 datasets} from the OpenML-CC18 data suite, up to 30,000 training samples. Larger sample sizes and more datasets could not be included due to MFs' space constraints.
    Positive and negative differences are averaged separately at each sample size, reducing biases.
    }
\label{fig:cc18_XForest_mf_fill_diff}
\end{minipage}
\end{figure}


HTs always have the lowest accuracy when training on the Splice dataset, which remains almost constant as sample size increases (Figure \ref{fig:select_acc_stream}, left). In this case, the classifier fails to learn any new information from streaming data batches.
Its performance also shows significant fluctuations when training on the other two datasets. 
The accuracy starts shifting sharply around 4,000 samples in the Pendigits task and around 32,000 samples in the CIFAR-10 task. 
We speculate that the fluctuations are due to how HT regulates low performing leaves under resource constraints \citep{domingos_mining_2000}. However, increasing the maximum size of HTs, even to excessive amounts over the memory limits, would not alleviate the problem. 
Thus, we attribute these fluctuations to possible implementation errors and mark the problematic sample sizes with the \textbf{cross symbol} (Figure \ref{fig:select_acc_stream}, center and right) instead of showing unreliable plots.
Such disadvantages could prevent HTs from producing meaningful results on certain datasets or maintaining consistent performance at larger sample sizes. Based on these results, we exclude forest ensembles of HTs from our benchmarks.

In the Pendigits task, MFs perform well at larger sample sizes (Figure \ref{fig:select_acc_stream}, center). However, their accuracy is worse than single XTrees' on the Splice dataset, only surpassing the almost constant plots of HTs (Figure \ref{fig:select_acc_stream}, left). 
We find MFs' worse performance in the Splice task as expected due to previous results on DNA data \citep{lakshminarayanan_mondrian_2014}. 
The accuracy could be improved if feature engineering is implemented, demonstrating the drawbacks of label-independent random splits on certain data domains \citep{roy_mondrian_2009, ziegler_mining_2014}.
Moreover, MFs perform worse than XForests on the CIFAR-10 dataset (Figure \ref{fig:select_acc_stream}, right). The results show that XForests are better at handling more feature dimensions.

\begin{figure*}[!htb]
\centering
\includegraphics[width=0.9\textwidth]{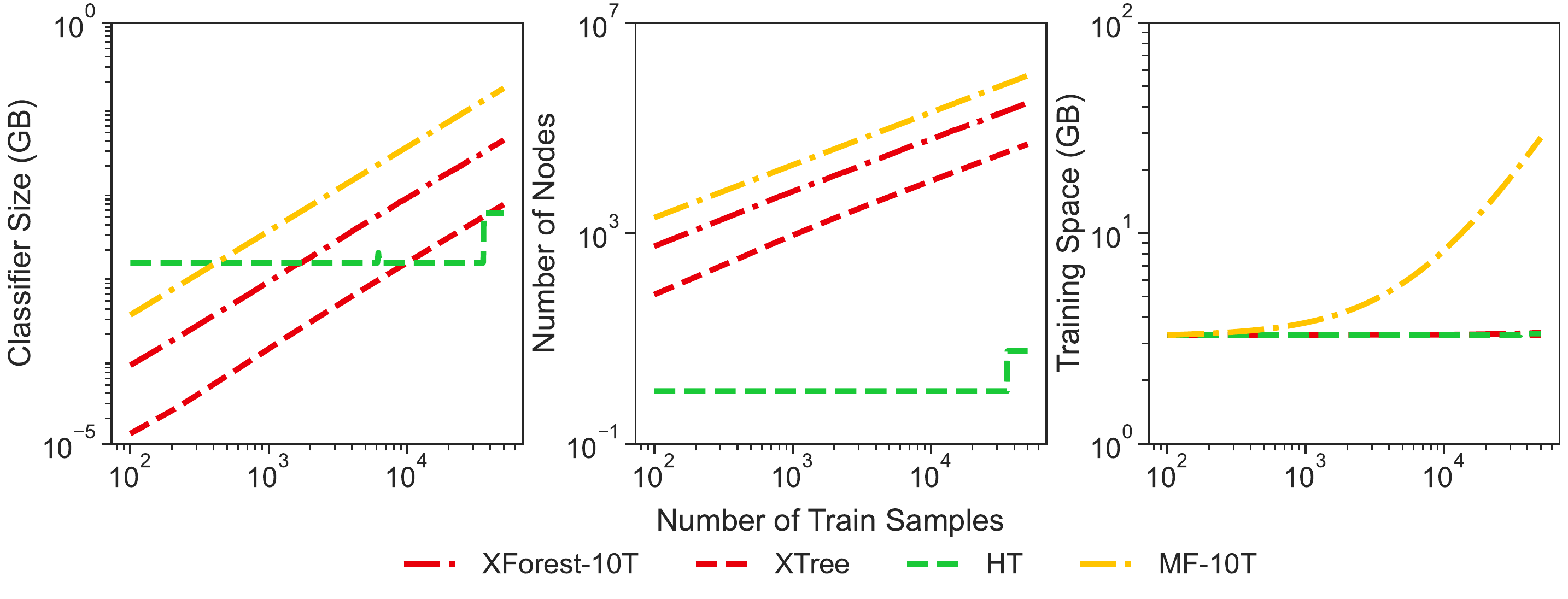}
  \caption{Streaming algorithm space complexities on the CIFAR-10 dataset. The classifiers include: Extremely Simple Streaming Forest with 10 trees \textbf{(XForest-10T)}, Extremely Simple Streaming Tree \textbf{(XTree)}, Hoeffding Tree \textbf{(HT)}, and Mondrian Forest with 10 trees \textbf{(MF-10T)}.
  Each line represents averaged results from \textbf{10 randomized repetitions}. Classifier size measures the pickled python objects, and training space checks the maximum virtual memories during fitting.
  XForests are the most efficient forest, and XTrees also show very low usage. 
  HTs' nodes fail to increase linearly with sample size, showing little improvement from new samples.
  MFs are the most space inefficient and take up excessive memories as sample size increases. 
  }
\label{fig:select_cifar_mem_stream}
\end{figure*}

Further comparisons on the OpenML-CC18 data suite (Figure \ref{fig:cc18_XForest_mf_fill_diff}) show that XForests and MFs have similar performance across different sample sizes. According to their signs, we average and plot the differences separately, minimizing the biases of particular datasets. Accuracy results use 5-fold cross validations on all available samples, and separate accuracy plots for each task are included in Appendix \ref{app:cc18}.

\subsection{Complexity Comparisons with Streaming Algorithms}
In terms of computational space, XForests are the most efficient forests on the CIFAR-10 dataset (Figure \ref{fig:select_cifar_mem_stream}). 
XTrees also show space efficiency by taking up small training space and low storage sizes.
However, the other two streaming classifiers: HTs and MFs both show abnormal space usage.
HTs' number of nodes remains constant until very large samples sizes (Figure \ref{fig:select_cifar_mem_stream}, center). Despite fitting new batches of training data, the classifier fails to generate new splits, and the results correspond to its suboptimal accuracy (Figure \ref{fig:select_acc_stream}, right).
Moreover, MFs take up excessive training space as sample size increases, substantially surpassing all other algorithms after 1,000 samples \ref{fig:select_cifar_mem_stream}, left and right). At the maximum size, they use six times more resources than XForests.
Such space constraints significantly limit the algorithm's applications. The total number of trees in MFs must be restricted to ensure task completion, limiting their potential performance. 
Thus, XForests would be considered the optimal choice for streaming tasks with limited resources. We do not report the other two tasks' virtual memories as the changes are minimal. 

See Appendix \ref{app:select_time} for training wall times on the three selected datasets.


\begin{figure}[!htb]
    \centering
    \includegraphics[width=0.8\columnwidth]{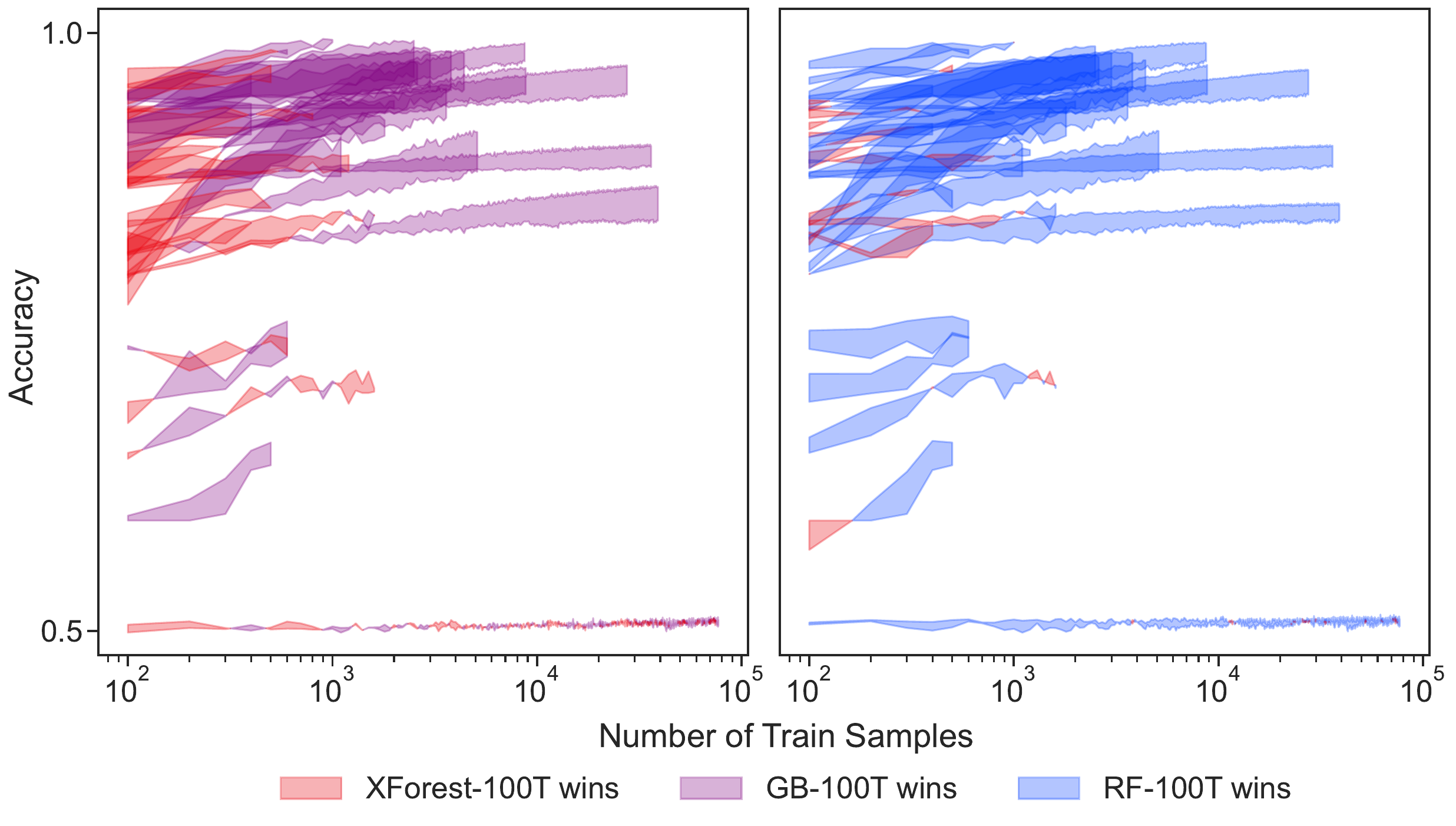}
    \caption{Accuracy with similar effect sizes (minimum $\geq -5\%$ and maximum $\leq 5\%$). The 100-tree classifiers include: Extremely Simple Streaming Forest\textbf{(XForest-100T)}, Gradient Boosting \textbf{(GB-100T)}, and Random Forest \textbf{(RF-100T)}.
    Each filled area represents average accuracy with \textbf{5-fold cross validations} from one of \textbf{30 (left)} or \textbf{37 (right)} classification tasks. 
    XForests would perform better than batch algorithms at sample sizes smaller than 2,000, and such accuracy similarities persist at larger sample sizes.
    }
\label{fig:cc18_fill}
\end{figure}


\subsection{Accuracy Comparisons with Batch Algorithms}
In the OpenML-CC18 tasks, XForests achieve high classification accuracy and consistent improvements as more data batches come in. They often achieve performance as well as, sometimes even better than, that of RFs and GBs (Appendix \ref{app:cc18}). These results illustrate that the ensemble can maintain its performance on datasets across a variety of data domains, including tabular, image, audio, and more \citep{bischl_openml_2019}. Such consistency makes XForests applicable to diverse real-world scenarios, avoiding the disadvantages of HTs and MFs.


We then evaluate performance similarities by calculating the effect sizes. Taken the pair of XForest and GB as an example, we average the classification accuracy across five folds and measure the difference ratios as:
\begin{equation*} \frac{\text{Average accuracy of XForest} - \text{Average accuracy of GB}}{\text{Average accuracy of GB}} \tag{2} \end{equation*}
Positive percentages represent the better performance of XForests, while negative values represent that of GBs. 
As sample size increases, 30 of the 72 tasks show very small shifts on effect sizes (Figure \ref{fig:cc18_fill}, left), each having a minimum $\geq -5\%$ and a maximum $\leq 5\%$. The same comparisons also apply to RFs (Figure \ref{fig:cc18_fill}, right) and include 37 tasks instead. 
On these datasets, XForests usually perform better than GBs and RFs at sample sizes smaller than 2,000, and they maintain similar accuracy at larger sizes. 
In the later situation, having more data available allows batch forests to establish finer tree structures and achieve better performance. 
Nevertheless, this scenario is unreasonable in real-world applications, and refitting batch forests at large sample sizes demands much more computational resources than incrementally updating XForests.

\begin{figure}[!htb]
\begin{minipage}[c]{0.45\columnwidth}
    \includegraphics[width=\columnwidth]{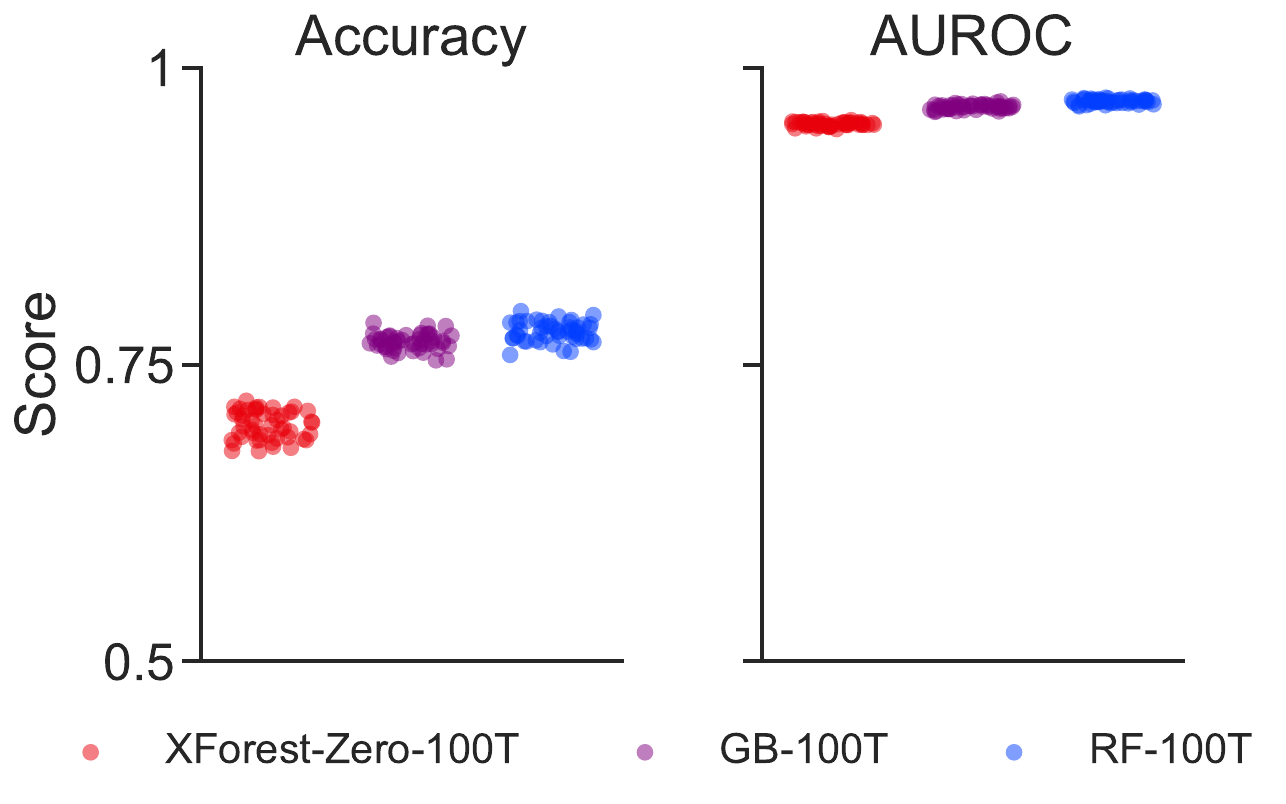}
\end{minipage}\hfill
\begin{minipage}[c]{0.5\columnwidth}
    \caption{Classifier accuracy and AUROC (one-vs-rest) on Fashion-MNIST test sets. The 100-tree classifiers include: Extremely Simple Streaming Forest \textbf{(XForest-Zero-100T)}, Gradient Boosting \textbf{(GB-100T)}, and Random Forest \textbf{(RF-100T)}. XForest-Zeros achieve very similar performance to decision forests only by refitting leaf nodes with Fashion-MNIST samples. Existing splits based on MNIST samples are efficiently applied to a new task.
    }
\label{fig:f_mnist_cat}
\end{minipage}
\end{figure}

In the MNIST-and-Fashion-MNIST task, XForest-Zeros perform competitively with RFs trained on Fashion-MNIST samples (Figure \ref{fig:f_mnist_cat}). Over 50 iterations, tree splits based on MNIST achieve consistent accuracy and AUROC on Fashion-MNIST by only refitting the posterior probabilities of leaf nodes. Compared to initializing and fitting a RF, an XForest-Zero only requires inference time and zero additional space. It efficiently utilizes the existing tree structures and transfer the splits to a similar yet new classification task.

In conclusion, we demonstrate that our algorithms, an extremely simple approach to streaming trees and ensembles, achieve consistently high performance on a variety of real-world datasets. 
We also illustrate that XForests avoid the current problems of HTs and MFs, which perform catastrophically on certain datasets and take up excessive memories at large sample sizes. Furthermore, XForest-Zeros show the applicability of efficiently transferring existing splits to another task.

\section{Theoretical Efficiency and Comparisons}
\label{theory}
We explore and compare the theoretical complexities of XForests and RFs. XForests are shown to be more efficient than RFs in learning new data. Theses comparisons show the effects of streaming batches on estimator complexities.

\subsection{Learning Time}
\label{theory:time}
\begin{lemma}
In terms of learning time complexity, XForests are more efficient than batch decision forests.
\end{lemma}
\begin{proof}
An algorithm's learning time complexity represents the theoretical learning time for a new dataset. It depends on both input properties and model hyperparameters. Let $T$ be the number of trees, $n$ the total number of training samples, $b$ the number of batches, $d$ the total number of features, and $p$ the number of features sampled at each split. For building RFs, the average time complexity is $\mathcal{O}(Tpn\log^2{n})$. The $pn\log{n}$ term corresponds to sorting $p$ features at each node, and $\log{n}$ accounts for the average number of nodes \citep{louppe_understanding_2015}. 

The average learning time of XForest only defers from RF's in terms of sample size. When training a new tree with one data batch, the average number of nodes is reduced to $\log(n/b)$, and sorting features in all nodes takes $\mathcal{O}(p(n/b)\log^2(n/b))$. Each time to update the tree with one new batch, XForest potentially visits all the existing nodes, taking a loose-bounded complexity of $\mathcal{O}(p(n/b)\log(n/b)\log(n))$. In early updates, the actual number of nodes visited would be much fewer than $\log(n)$. When considering all the batches, the updating complexity becomes:
\begin{align*}
    \mathcal{O}(bp(n/b)\log(n/b)\log(n))
    &= \mathcal{O}(pn\log(n/b)\log(n))
\end{align*}

At the forest level, fully building the estimator includes additions and updates for every single tree: 
\begin{align*}
    \mathcal{O}(T(p(n/b)\log^2{(n/b)} + pn\log(n/b)\log(n)))
\end{align*}

As the updating process is slower, the complexity can be simplified as $\mathcal{O}(Tpn\log(n/b)\log(n))$, which is faster than building RFs when $b \geq 2$.
\end{proof}

\subsection{Learning Space}
\begin{lemma}
In terms of learning space complexity, XForests are more efficient than batch decision forests.
\end{lemma} \vspace{-0.3cm}
\begin{proof}
An algorithm's learning space complexity describes the theoretical maximum computational space during training, which scales with sample size and hyperparameters. Let $c$ be the number of classes and $T$, $d$, $n$, and $b$ be defined as in \ref{theory:time}. 

Building RFs requires keeping all the samples in memory, so the data matrix requires a complexity of $\mathcal{O}(dn)$. When splitting nodes, two $c$-length arrays record the class counts on either end of the split point. They are also used to evaluate the splitting criteria, such as Gini impurity and entropy. Fully grown trees have totally $2n - 1$ nodes, which contain exactly one sample in every leaf. This term is dominated by the $dn$ term, making RF's space complexity $\mathcal{O}(dn + c)$.

XForest updates the tree structure with streaming data batches, discarding the old data after each fitting (Algorithm \ref{alg:XForest}). Data storage only accounts for the current batch and has an average complexity of $\mathcal{O}(dn/b)$. The class counts remain the same, and fully grown trees also have the same maximum. Thus, XForest's learning space complexity becomes $\mathcal{O}(dn/b + c)$, which is more efficient than RF's.
\end{proof}



\section{Discussion}
\label{discussion}
Our streaming tree algorithms: XTree and XForest utilize an extremely simple strategy of extending fitted decision trees (Algorithm \ref{alg:XTree} and \ref{alg:XForest}). It maintains all the existing partitions and further splits the feature space with new data batches (Figure \ref{fig:demo}). This approach is easier to implement than Hoeffding bounds and Mondrian processes, as both methods modify the tree splitting mechanisms \citep{hoeffding_probability_1994, domingos_mining_2000, roy_mondrian_2009, lakshminarayanan_mondrian_2014}. 
Furthermore, XForests achieve consistently high accuracy in diverse classification tasks and often perform as well as, occasionally even better than, state-of-the-art forests (Figure \ref{fig:select_acc_stream}, \ref{fig:cc18_XForest_mf_fill_diff} and \ref{fig:cc18}). It is also more time and space efficient for XForests to learn new information (Section \ref{theory}).
These results show that our methods avoid the problems of HTs and MFs, including low accuracy in some tasks and excessive memories in others (Figure \ref{fig:select_acc_stream} and \ref{fig:select_cifar_mem_stream}). Thus, XTrees and XForests establish a simple standard for streaming trees that could be readily applied to many real-world problems.

On some datasets with high complexities, batch trees and ensembles could achieve more accurate results if all samples are given at once (Figure \ref{fig:cc18_fill} and \ref{fig:cc18}). Nonetheless, this assumption is unreasonable in many real-world scenarios \citep{abdulsalam_streaming_2007, liu_isolation_2008}. In our experiments and actual practices, batch decision trees and forests need to store all available data and refit them every time a new batch of samples comes in. Even by restricting the number of refitting with certain criteria (e.g. a minimum amount of new samples), it would still cost significantly more computational resources than incremental updates (Section \ref{theory}). Thus, our streaming trees provide an alternative solution to such tasks. As an XForest continuously improves its partitions with new information, any differences in performance would eventually be eliminated by enough data batches. It is therefore more economical to keep updating an established streaming model than constructing a new batch forest from time to time.

While our XForest experiments focus on single in-distribution tasks, its applications could be extended to solve OOD problems like distribution shift and continual learning \citep{geisa_towards_2021, van_de_ven_three_2019}. 
Technically, each extension of decision trees does not have to rely on samples from the same distribution or use the same splitting criteria. Old trees can also be replaced anytime if previous partitions fail to perform well on current data streams. Moreover, as supplemental data arrive, continual learning models could leverage XForests to incrementally update existing tasks (Appendix \ref{app:xor_xnor} and \ref{app:xor_rxor}).
Finally, XForest-Zeros also exhibit another method of utilizing existing tree structures across different tasks, which even has higher efficiency.

Overall, our method of incremental updating and refitting excels at simplicity, consistency, learning time and space efficiency, and extensibility. This approach could offer great development potentials and set a clear standard for streaming tree algorithms, focusing on real-world applications.

\section*{Acknowledgements}
The authors acknowledge the National Science Foundation-Simons Foundation’s Research Collaboration on the Mathematical and Scientific Foundations of Deep Learning (MoDL), NSF grant 2031985. We also thank Vivek Gopalakrishnan, Michael Powell, Tyler Tomita, Randal Burns, Meghana Madhyastha, Nicholas Hahn, and all members of the NeuroData Lab at JHU for their valuable supports. This work is graciously funded by the Defense Advanced Research Projects Agency (DARPA) Lifelong Learning Machines program through contract FA8650-18-2-7834. Experiments were partially supported by funding from Microsoft Research.

\bibliography{ref}

\begin{thebibliography}{44}
\providecommand{\natexlab}[1]{#1}
\providecommand{\url}[1]{\texttt{#1}}
\expandafter\ifx\csname urlstyle\endcsname\relax
  \providecommand{\doi}[1]{doi: #1}\else
  \providecommand{\doi}{doi: \begingroup \urlstyle{rm}\Url}\fi

\bibitem[Abdulsalam et~al.(2007)Abdulsalam, Skillicorn, and Martin]{abdulsalam_streaming_2007}
Hanady Abdulsalam, David~B. Skillicorn, and Patrick Martin.
\newblock Streaming {Random} {Forests}.
\newblock In \emph{11th {International} {Database} {Engineering} and {Applications} {Symposium} ({IDEAS} 2007)}, pp.\  225--232, 2007.

\bibitem[Alimoglu \& Alpaydin(1996)Alimoglu and Alpaydin]{alimoglu_methods_1996}
Fevzi Alimoglu and Ethem Alpaydin.
\newblock Methods of {Combining} {Multiple} {Classifiers} {Based} on {Different} {Representations} for {Pen}-based {Handwritten} {Digit} {Recognition}.
\newblock In \emph{Proceedings of the {Fifth} {Turkish} {Artificial} {Intelligence} and {Artificial} {Neural} {Networks} {Symposium} ({TAINN} 96}, 1996.

\bibitem[Alpaydin \& Alimoglu(1998)Alpaydin and Alimoglu]{alpaydin_pen-based_1998}
E.~Alpaydin and Fevzi. Alimoglu.
\newblock Pen-{Based} {Recognition} of {Handwritten} {Digits}, 1998.

\bibitem[Amit \& Geman(1997)Amit and Geman]{amit_shape_1997}
Yali Amit and Donald Geman.
\newblock Shape {Quantization} and {Recognition} with {Randomized} {Trees}.
\newblock \emph{Neural Computation}, 9\penalty0 (7):\penalty0 1545--1588, July 1997.

\bibitem[Athey et~al.(2018)Athey, Tibshirani, and Wager]{athey_generalized_2018}
Susan Athey, Julie Tibshirani, and Stefan Wager.
\newblock Generalized {Random} {Forests}.
\newblock arXiv:1610.01271 [stat], April 2018.
\newblock URL \url{http://arxiv.org/abs/1610.01271}.

\bibitem[Ben-Haim \& Tom-Tov(2010)Ben-Haim and Tom-Tov]{ben-haim_streaming_2010}
Yael Ben-Haim and Elad Tom-Tov.
\newblock A {Streaming} {Parallel} {Decision} {Tree} {Algorithm}.
\newblock \emph{Journal of Machine Learning Research}, 11\penalty0 (Feb):\penalty0 849--872, 2010.

\bibitem[Biau et~al.(2008)Biau, Devroye, and Lugosi]{biau_consistency_2008}
G{\'e}rard Biau, Luc Devroye, and G{\'a}bor Lugosi.
\newblock Consistency of {Random} {Forests} and {Other} {Averaging} {Classifiers}.
\newblock \emph{Journal of Machine Learning Research}, 9\penalty0 (66):\penalty0 2015--2033, 2008.

\bibitem[Bifet \& Gavalda(2009)Bifet and Gavalda]{bifet_adaptive_2009}
Albert Bifet and Ricard Gavalda.
\newblock Adaptive learning from evolving data streams.
\newblock In \emph{International {Symposium} on {Intelligent} {Data} {Analysis}}, pp.\  249--260. Springer, 2009.

\bibitem[Bischl et~al.(2019)Bischl, Casalicchio, Feurer, Hutter, Lang, Mantovani, Rijn, and Vanschoren]{bischl_openml_2019}
Bernd Bischl, Giuseppe Casalicchio, Matthias Feurer, Frank Hutter, Michel Lang, Rafael~G. Mantovani, Jan N.~van Rijn, and Joaquin Vanschoren.
\newblock {OpenML} {Benchmarking} {Suites}.
\newblock 2019.

\bibitem[Breiman(2001)]{breiman_random_2001}
Leo Breiman.
\newblock Random {Forests}.
\newblock \emph{Machine Learning}, 45\penalty0 (1):\penalty0 5--32, 2001.

\bibitem[Caruana \& Niculescu-Mizil(2006)Caruana and Niculescu-Mizil]{caruana_empirical_2006}
Rich Caruana and Alexandru Niculescu-Mizil.
\newblock An {Empirical} {Comparison} of {Supervised} {Learning} {Algorithms}.
\newblock In \emph{Proceedings of the 23rd {International} {Conference} on {Machine} {Learning}}, {ICML} '06, pp.\  161--168, New York, NY, USA, 2006. ACM.

\bibitem[Caruana et~al.(2008)Caruana, Karampatziakis, and Yessenalina]{caruana_empirical_2008}
Rich Caruana, Nikos Karampatziakis, and Ainur Yessenalina.
\newblock An empirical evaluation of supervised learning in high dimensions.
\newblock In \emph{Proceedings of the 25th international conference on {Machine} learning}, pp.\  96--103, New York, New York, USA, July 2008. ACM.

\bibitem[Chen \& Guestrin(2016)Chen and Guestrin]{chen_xgboost_2016}
Tianqi Chen and Carlos Guestrin.
\newblock {XGBoost}: {A} {Scalable} {Tree} {Boosting} {System}.
\newblock In \emph{Proceedings of the 22nd {ACM} {SIGKDD} {International} {Conference} on {Knowledge} {Discovery} and {Data} {Mining}}, {KDD} '16, pp.\  785--794, New York, NY, USA, August 2016. Association for Computing Machinery.

\bibitem[Dey et~al.(2024)Dey, Geisa, Mehta, Tomita, Helm, Xu, Eaton, Dick, Priebe, and Vogelstein]{geisa_towards_2021}
Jayanta Dey, Ali Geisa, Ronak Mehta, Tyler~M. Tomita, Hayden~S. Helm, Haoyin Xu, Eric Eaton, Jeffery Dick, Carey~E. Priebe, and Joshua~T. Vogelstein.
\newblock Towards a theory of out-of-distribution learning.
\newblock arXiv:2109.14501 [stat], June 2024.
\newblock URL \url{http://arxiv.org/abs/2109.14501}.

\bibitem[Domingos \& Hulten(2000)Domingos and Hulten]{domingos_mining_2000}
Pedro Domingos and Geoff Hulten.
\newblock Mining {High}-{Speed} {Data} {Streams}.
\newblock In \emph{Proceedings of the {Sixth} {ACM} {SIGKDD} {International} {Conference} on {Knowledge} {Discovery} and {Data} {Mining}}, {KDD} '00, pp.\  71--80, New York, NY, USA, 2000. Association for Computing Machinery.

\bibitem[Fern{\'a}ndez-Delgado et~al.(2014)Fern{\'a}ndez-Delgado, Cernadas, Barro, and Amorim]{fernandez-delgado_we_2014}
Manuel Fern{\'a}ndez-Delgado, Eva Cernadas, Sen{\'e}n Barro, and Dinani Amorim.
\newblock Do we {Need} {Hundreds} of {Classifiers} to {Solve} {Real} {World} {Classification} {Problems}?
\newblock \emph{Journal of Machine Learning Research}, 15\penalty0 (90):\penalty0 3133--3181, 2014.

\bibitem[Feurer et~al.(2019)Feurer, Rijn, Kadra, Gijsbers, Mallik, Ravi, Mueller, Vanschoren, and Hutter]{feurer_openml-python_2019}
Matthias Feurer, Jan N.~van Rijn, Arlind Kadra, Pieter Gijsbers, Neeratyoy Mallik, Sahithya Ravi, Andreas Mueller, Joaquin Vanschoren, and Frank Hutter.
\newblock {OpenML}-{Python}: an extensible {Python} {API} for {OpenML}.
\newblock 2019.

\bibitem[Friedman(2001)]{friedman_greedy_2001}
Jerome~H. Friedman.
\newblock Greedy function approximation: {A} gradient boosting machine.
\newblock \emph{The Annals of Statistics}, 29\penalty0 (5):\penalty0 1189--1232, October 2001.

\bibitem[Gomes et~al.(2019)Gomes, Read, Bifet, Barddal, and Gama]{gomes_machine_2019}
Heitor~Murilo Gomes, Jesse Read, Albert Bifet, Jean~Paul Barddal, and Jo{\~a}o Gama.
\newblock Machine learning for streaming data: state of the art, challenges, and opportunities.
\newblock \emph{ACM SIGKDD Explorations Newsletter}, 21\penalty0 (2):\penalty0 6--22, November 2019.

\bibitem[Hoeffding(1994)]{hoeffding_probability_1994}
Wassily Hoeffding.
\newblock Probability {Inequalities} for sums of {Bounded} {Random} {Variables}.
\newblock In N.~I. Fisher and P.~K. Sen (eds.), \emph{The {Collected} {Works} of {Wassily} {Hoeffding}}, Springer {Series} in {Statistics}, pp.\  409--426. Springer, New York, NY, 1994.

\bibitem[{Joblib developers}(2022)]{joblib_developers_joblibjoblib_2022}
{Joblib developers}.
\newblock joblib/joblib, January 2022.

\bibitem[Jordan \& Mitchell(2015)Jordan and Mitchell]{jordan_machine_2015}
M.~I. Jordan and T.~M. Mitchell.
\newblock Machine learning: {Trends}, perspectives, and prospects.
\newblock \emph{Science}, 349\penalty0 (6245):\penalty0 255--260, 2015.

\bibitem[Khannouz \& Glatard(2020)Khannouz and Glatard]{khannouz_benchmark_2020}
Martin Khannouz and Tristan Glatard.
\newblock A {Benchmark} of {Data} {Stream} {Classification} for {Human} {Activity} {Recognition} on {Connected} {Objects}.
\newblock \emph{Sensors}, 20\penalty0 (22), 2020.

\bibitem[Krizhevsky(2012)]{krizhevsky_learning_2012}
Alex Krizhevsky.
\newblock Learning {Multiple} {Layers} of {Features} from {Tiny} {Images}.
\newblock \emph{University of Toronto}, 2012.

\bibitem[Kumar(2017)]{kumar_scikit-gardenscikit-garden_2017}
M~Kumar.
\newblock Scikit-garden/scikit-garden: {A} garden for scikit-learn compatible trees.
\newblock \emph{Accessed: Apr}, 13:\penalty0 2018, 2017.

\bibitem[Lakshminarayanan et~al.(2014)Lakshminarayanan, Roy, and Teh]{lakshminarayanan_mondrian_2014}
Balaji Lakshminarayanan, Daniel~M Roy, and Yee~Whye Teh.
\newblock Mondrian {Forests}: {Efficient} {Online} {Random} {Forests}.
\newblock In Z.~Ghahramani, M.~Welling, C.~Cortes, N.~Lawrence, and K.~Q. Weinberger (eds.), \emph{Advances in {Neural} {Information} {Processing} {Systems}}, volume~27. Curran Associates, Inc., 2014.

\bibitem[Lakshminarayanan et~al.(2016)Lakshminarayanan, Roy, and Teh]{lakshminarayanan_mondrian_2016}
Balaji Lakshminarayanan, Daniel~M. Roy, and Yee~Whye Teh.
\newblock Mondrian {Forests} for {Large}-{Scale} {Regression} when {Uncertainty} {Matters}.
\newblock In \emph{Proceedings of the 19th {International} {Conference} on {Artificial} {Intelligence} and {Statistics}}, pp.\  1478--1487. PMLR, May 2016.

\bibitem[Lavanya et~al.(2017)Lavanya, Bajaj, Tank, and Jain]{lavanya_handwritten_2017}
K~Lavanya, Shaurya Bajaj, Prateek Tank, and Shashwat Jain.
\newblock Handwritten digit recognition using hoeffding tree, decision tree and random forests --- {A} comparative approach.
\newblock In \emph{2017 {International} {Conference} on {Computational} {Intelligence} in {Data} {Science}({ICCIDS})}, pp.\  1--6, June 2017.

\bibitem[Lecun et~al.(1998)Lecun, Bottou, Bengio, and Haffner]{lecun_gradient-based_1998}
Y.~Lecun, L.~Bottou, Y.~Bengio, and P.~Haffner.
\newblock Gradient-based learning applied to document recognition.
\newblock \emph{Proceedings of the IEEE}, 86\penalty0 (11):\penalty0 2278--2324, November 1998.
\newblock ISSN 1558-2256.
\newblock \doi{10.1109/5.726791}.
\newblock URL \url{https://ieeexplore.ieee.org/document/726791}.
\newblock Conference Name: Proceedings of the IEEE.

\bibitem[Liaw et~al.(2002)Liaw, Wiener, and {others}]{liaw_classification_2002}
Andy Liaw, Matthew Wiener, and {others}.
\newblock Classification and regression by {randomForest}.
\newblock \emph{R news}, 2\penalty0 (3):\penalty0 18--22, 2002.

\bibitem[Liu et~al.(2008)Liu, Ting, and Zhou]{liu_isolation_2008}
Fei~Tony Liu, Kai~Ming Ting, and Zhi-Hua Zhou.
\newblock Isolation {Forest}.
\newblock In \emph{2008 {Eighth} {IEEE} {International} {Conference} on {Data} {Mining}}, pp.\  413--422, December 2008.

\bibitem[Louppe(2015)]{louppe_understanding_2015}
Gilles Louppe.
\newblock Understanding {Random} {Forests}: {From} {Theory} to {Practice}.
\newblock \emph{arXiv:1407.7502 [stat]}, June 2015.

\bibitem[Montiel et~al.(2020)Montiel, Halford, Mastelini, Bolmier, Sourty, Vaysse, Zouitine, Gomes, Read, Abdessalem, and Bifet]{montiel_river_2020}
Jacob Montiel, Max Halford, Saulo~Martiello Mastelini, Geoffrey Bolmier, Raphael Sourty, Robin Vaysse, Adil Zouitine, Heitor~Murilo Gomes, Jesse Read, Talel Abdessalem, and Albert Bifet.
\newblock River: machine learning for streaming data in {Python}, 2020.

\bibitem[Pedregosa et~al.(2011)Pedregosa, Varoquaux, Gramfort, Michel, Thirion, Grisel, Blondel, Prettenhofer, Weiss, Dubourg, Vanderplas, Passos, Cournapeau, Brucher, Perrot, and Duchesnay]{pedregosa_scikit-learn_2011}
F.~Pedregosa, G.~Varoquaux, A.~Gramfort, V.~Michel, B.~Thirion, O.~Grisel, M.~Blondel, P.~Prettenhofer, R.~Weiss, V.~Dubourg, J.~Vanderplas, A.~Passos, D.~Cournapeau, M.~Brucher, M.~Perrot, and E.~Duchesnay.
\newblock Scikit-learn: {Machine} {Learning} in {Python}.
\newblock \emph{Journal of Machine Learning Research}, 12:\penalty0 2825--2830, 2011.

\bibitem[Pfahringer et~al.(2007)Pfahringer, Holmes, and Kirkby]{pfahringer_new_2007}
Bernhard Pfahringer, Geoffrey Holmes, and Richard Kirkby.
\newblock New {Options} for {Hoeffding} {Trees}.
\newblock In Mehmet~A. Orgun and John Thornton (eds.), \emph{{AI} 2007: {Advances} in {Artificial} {Intelligence}}, Lecture {Notes} in {Computer} {Science}, pp.\  90--99, Berlin, Heidelberg, 2007. Springer.

\bibitem[Rampone(1998)]{rampone_splice-junction_1998}
S.~Rampone.
\newblock Splice-junction recognition on gene sequences ({DNA}) by {BRAIN} learning algorithm.
\newblock In \emph{1998 {IEEE} {International} {Joint} {Conference} on {Neural} {Networks} {Proceedings}. {IEEE} {World} {Congress} on {Computational} {Intelligence} ({Cat}. {No}.{98CH36227})}, volume~1, pp.\  774--779 vol.1, 1998.

\bibitem[Rodola(2022)]{rodola_giampaolopsutil_2022}
Giampaolo Rodola.
\newblock giampaolo/psutil, January 2022.

\bibitem[Roy \& Teh(2009)Roy and Teh]{roy_mondrian_2009}
DM~Roy and YW~Teh.
\newblock The {Mondrian} process.
\newblock In \emph{Advances in {Neural} {Information} {Processing} {Systems}}, 2009.

\bibitem[Sarkar et~al.(2020)Sarkar, Chatterjee, Das, and Mondal]{sarkar_splice_2020}
Rahul Sarkar, Chandra~Churh Chatterjee, Sayantan Das, and Dhiman Mondal.
\newblock Splice {Junction} {Prediction} in {DNA} {Sequence} {Using} {Multilayered} {RNN} {Model}.
\newblock In Suresh~Chandra Satapathy, K.~Srujan Raju, K.~Shyamala, D.~Rama Krishna, and Margarita~N. Favorskaya (eds.), \emph{Advances in {Decision} {Sciences}, {Image} {Processing}, {Security} and {Computer} {Vision}}, pp.\  39--47, Cham, 2020. Springer International Publishing.

\bibitem[Towell et~al.(1991)Towell, Noordewier, and Shavlik]{towell_molecular_1991}
G.~Towell, M.~Noordewier, and J.~Shavlik.
\newblock Molecular {Biology} ({Splice}-junction {Gene} {Sequences}), 1991.

\bibitem[van~de Ven \& Tolias(2019)van~de Ven and Tolias]{van_de_ven_three_2019}
Gido~M. van~de Ven and Andreas~S. Tolias.
\newblock Three scenarios for continual learning.
\newblock \emph{arXiv:1904.07734 [cs, stat]}, April 2019.

\bibitem[Vanschoren et~al.(2013)Vanschoren, van Rijn, Bischl, and Torgo]{vanschoren_openml_2013}
Joaquin Vanschoren, Jan~N. van Rijn, Bernd Bischl, and Luis Torgo.
\newblock {OpenML}: {Networked} {Science} in {Machine} {Learning}.
\newblock \emph{SIGKDD Explorations}, 15\penalty0 (2):\penalty0 49--60, 2013.

\bibitem[Xiao et~al.(2017)Xiao, Rasul, and Vollgraf]{xiao_fashion-mnist_2017}
Han Xiao, Kashif Rasul, and Roland Vollgraf.
\newblock Fashion-{MNIST}: a {Novel} {Image} {Dataset} for {Benchmarking} {Machine} {Learning} {Algorithms}.
\newblock arXiv:1708.07747 [cs], September 2017.
\newblock URL \url{http://arxiv.org/abs/1708.07747}.

\bibitem[Ziegler \& K{\"o}nig(2014)Ziegler and K{\"o}nig]{ziegler_mining_2014}
Andreas Ziegler and Inke~R. K{\"o}nig.
\newblock Mining data with random forests: current options for real-world applications.
\newblock \emph{WIREs Data Mining and Knowledge Discovery}, 4\penalty0 (1):\penalty0 55--63, 2014.

\end{thebibliography}
\bibliographystyle{collas2025_conference}
\clearpage
\appendix
\section{Accuracy for OpenML-CC18 Tasks}
\label{app:cc18}

\begin{figure}[!htpb]
\centering
\includegraphics[width=0.8\columnwidth]{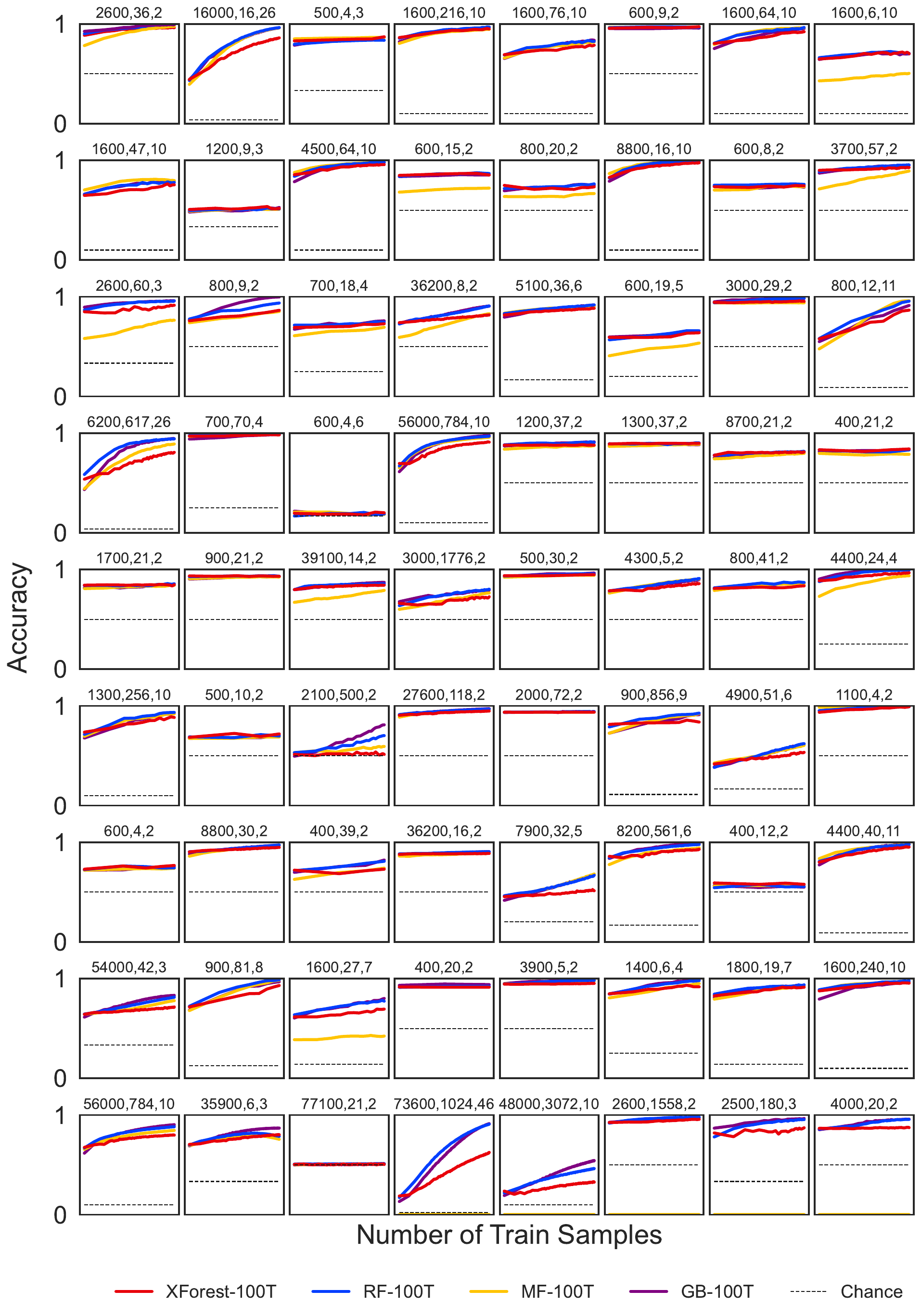}
  \caption{Classifications on the OpenML-CC18 datasets with max sample size, number of features, and number of classes. The 100-tree classifiers include: Extremely Simple Streaming Forest \textbf{(XForest-100T)}, Mondrian Forest \textbf{(MF-100T)}, Random Forest \textbf{(RF-100T)} and Gradient Boosting \textbf{(GB-100T)}. 
  All plots show averaged accuracy over five folds and are listed in the order of dataset IDs. Sample sizes correspond to respective datasets. In many tasks, XForests exceed MFs and perform as well as, sometimes even better than, RFs and GBs. Moreover, XForest accuracy consistently increases with new samples across different data domains.
  }
\label{fig:cc18}
\end{figure}

\clearpage

\section{Wall Times for Three Selected Tasks}
\label{app:select_time}

\begin{figure*}[!htb]
\centering
\includegraphics[width=0.9\textwidth]{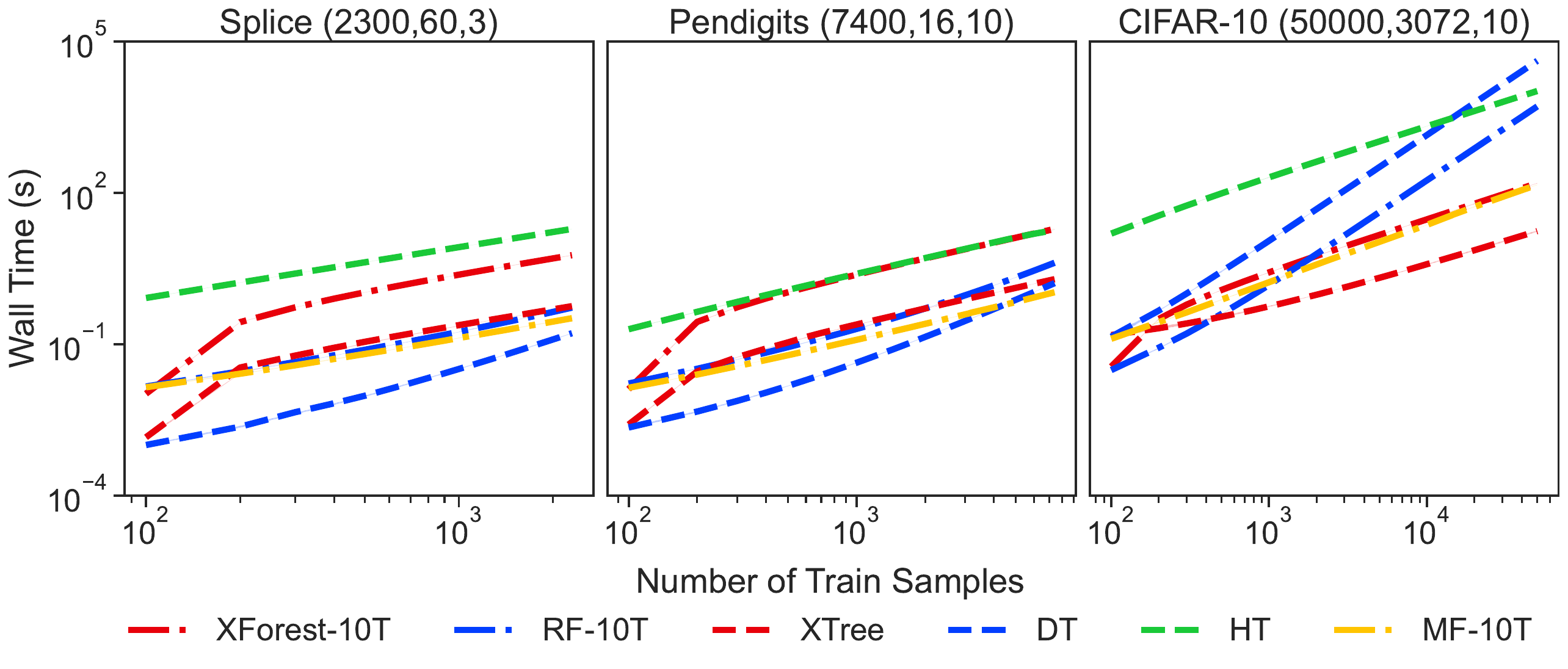}
  \caption{Training wall times on the Splice \textbf{(left)}, Pendigits \textbf{(center)}, and CIFAR-10 \textbf{(right)} datasets. The panel titles include the maximum sample size, the number of features, and the number of classes.
  The classifiers include: Extremely Simple Streaming Forest with 10 trees \textbf{(XForest-10T)}, Random Forest with 10 trees \textbf{(RF-10T)}, Extremely Simple Streaming Tree \textbf{(XTree)}, Decision Tree \textbf{(DT)}, Hoeffding Tree \textbf{(HT)}, and Mondrian Forest with 10 trees \textbf{(MF-10T)}. 
  Each line represents averaged results from \textbf{10 randomized repetitions}.
  Cumulative fitting times are shown for all estimators, and two batch estimators' times include refitting at each sample size. XTrees are the most efficient in the CIFAR-10 task. XForests take longer times in the Splice and Pendigits tasks, but are overtaken by HTs, batch DTs, and RFs in the CIFAR-10 task. MFs maintain similar time efficiency as XForests. RFs are faster than DTs in the CIFAR-10 task, which could be attributed to the feature reductions.
  }
\label{fig:select_time}
\end{figure*}

In terms of training wall times, XTrees are the most efficient on the CIFAR-10 dataset (Figure \ref{fig:select_time}, right). 
XForests take longer times in the Splice and Pendigits tasks (Figure \ref{fig:select_time}, left and center). However, on the CIFAR-10 dataset, HTs are the most inefficient streaming trees, even slower than the two streaming ensembles. Also, as sample size increases, time usage of batch estimators grows more quickly due to refitting, and they surpass XForests around 2,000 samples. MFs take similar training times as XForests.
All training wall times grow linearly as sample size increases. Results are subject to implementation details of packages used (Section \ref{methods}).

\clearpage

\section{XOR and XNOR Tasks}
\label{app:xor_xnor}

\begin{figure*}[!htb]
\centering
\includegraphics[width=0.9\textwidth]{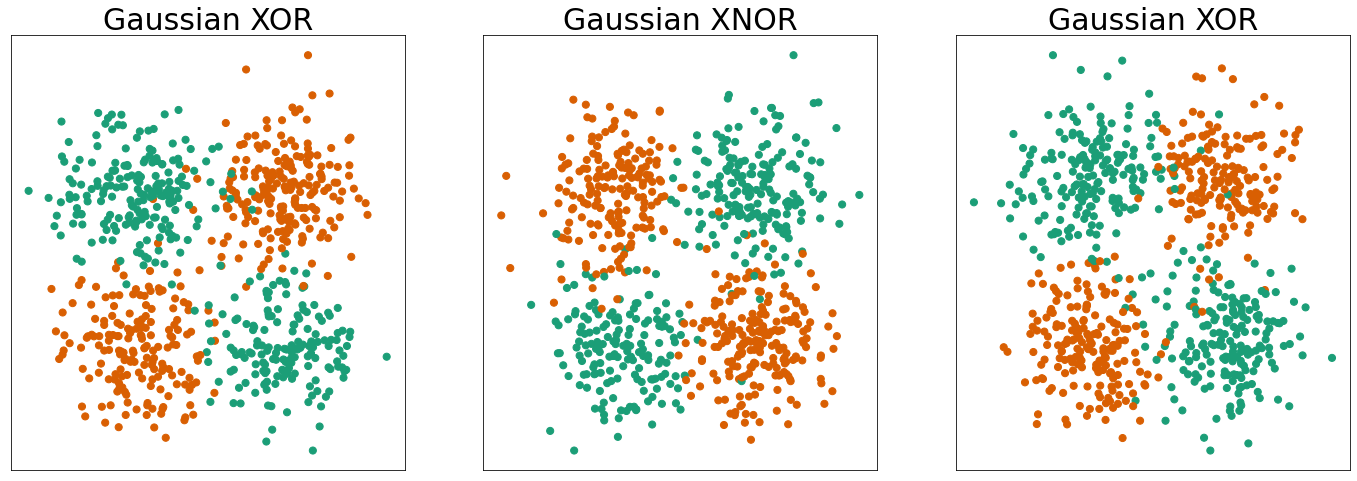}

\includegraphics[width=0.9\textwidth]{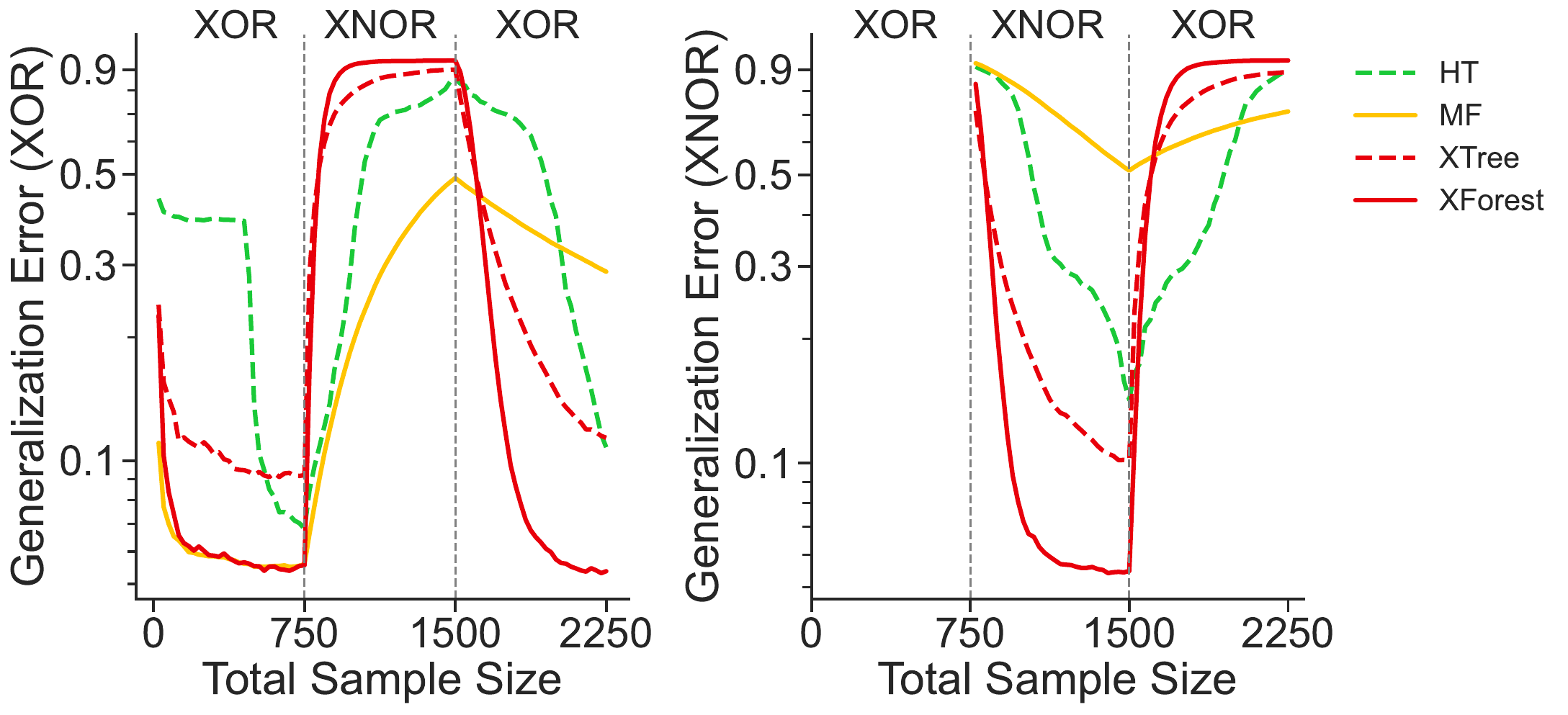}
  \caption{Distribution shift adaptation with Gaussian XOR and XNOR. XForests and XTrees adapt quickly to the new distribution after both shifts. 
  }
\label{fig:xor_xnor}
\end{figure*}

Figure \ref{fig:xor_xnor} shows XForests and XTrees can adapt to two distribution shifts in data batches. Top row shows 3 subsets of 750 samples belonging to two types of distributions. Gaussian XNOR has the same distribution as Gaussian XOR but with the inverse class labels. Data are introduced in batches of 25 samples at a time, and the distribution shifts at 750 samples and 1,500 samples. HTs have slower adaptation, while MFs fail to achieve good performance after both shifts.

\clearpage

\section{XOR and R-XOR Tasks}
\label{app:xor_rxor}

\begin{figure*}[!htb]
\centering
\includegraphics[width=0.9\textwidth]{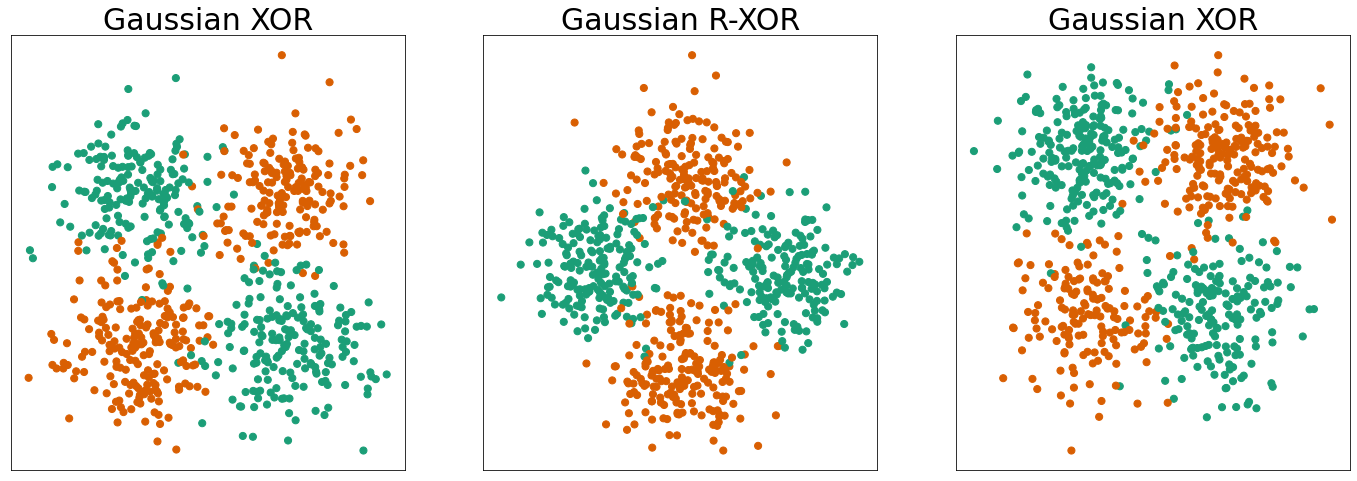}
\includegraphics[width=0.9\textwidth]{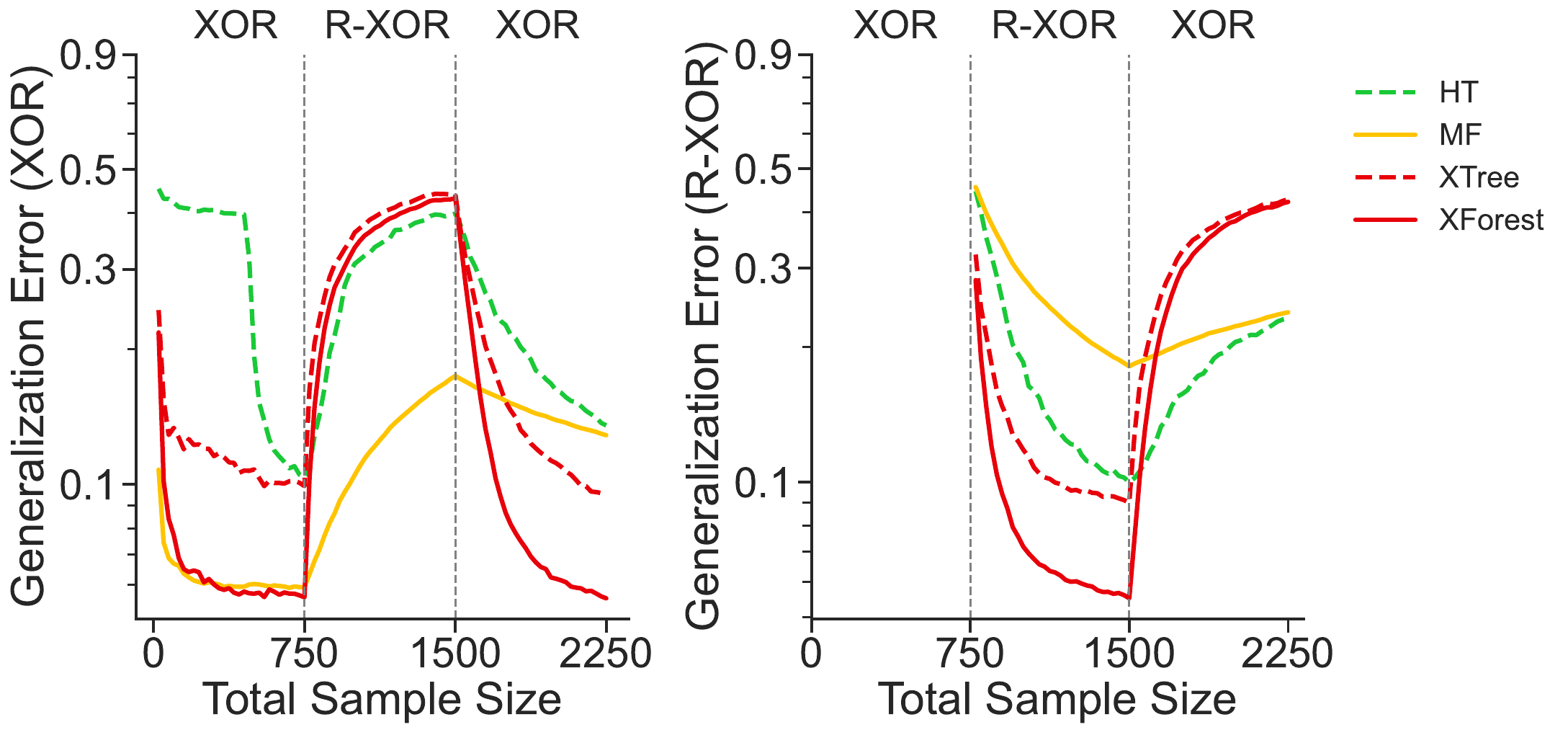}
  \caption{Distribution shift adaptation with Gaussian XOR and R-XOR. XForests and XTrees adapt quickly to the new distribution after both shifts. 
  }
\label{fig:xor_rxor}
\end{figure*}

Figure \ref{fig:xor_rxor} shows XForests and XTrees can adapt to two distribution shifts in data batches. Top row shows 3 subsets of 750 samples belonging to two types of distributions. Gaussian R-XOR has the same distribution as Gaussian XOR but with the class labels rotated 45 degrees. Data are introduced in batches of 25 samples at a time, and the distribution shifts at 750 samples and 1,500 samples. HTs have slower adaptation, while MFs fail to achieve good performance after both shifts.
\end{document}